%% file: main.tex
\documentclass{article}

\pdftrailerid{}

\PassOptionsToPackage{numbers,sort&compress}{natbib}
\usepackage[final]{neurips_2026}
\workshoptitle{AXIOM: Foundations of Efficient Deep Learning}

\usepackage[utf8]{inputenc}
\usepackage[T1]{fontenc}

\usepackage{amsmath}
\usepackage{amssymb}
\usepackage{booktabs}
\usepackage{graphicx}
\usepackage{microtype}
\usepackage{xcolor}
\usepackage{url}
\usepackage[hidelinks]{hyperref}
\hypersetup{pdfinfo={Author={},Subject={},Title={},Keywords={},Creator={},Producer={}}}

\title{Router Sensitivity Under Lightweight Fine-Tuning Identifies
Prunable Experts in Mixture-of-Experts Models}

\author{%
Ali Janati\textsuperscript{1,2,*}\quad
Kaoutar El Maghraoui\textsuperscript{2}\quad
Chengke Zou\textsuperscript{2} \\
\textbf{Xinyi Luo\textsuperscript{2}\quad Wenyuan Shen\textsuperscript{2}} \\
{\small \textsuperscript{1}Data Science Institute, Columbia University, New York, NY, USA} \\
{\small \textsuperscript{2}Department of Computer Science, Columbia University, New York, NY, USA} \\
}

\input{camera_ready_layout}

\begin{document}

\maketitle

\begin{abstract}
Mixture-of-Experts (MoE) models decouple parameter count from per-token compute, but deployment still requires hosting every expert in memory. Recent theory shows that experts whose router weights change least during fine-tuning can be pruned with provable accuracy preservation, yet the guarantee assumes full fine-tuning. We show that the signal can be elicited through a brief parameter-efficient adaptation. We fine-tune with a lightweight adapter, rank experts by the induced router change, and prune the least-changed experts in one shot. On Mixtral-8$\times$7B-Instruct, router-only LoRA trains 0.002\% of parameters and retains 27.54\% MMLU-Pro accuracy with half the experts removed, against roughly 16\% for magnitude and random pruning. Signal quality improves monotonically with adapter size, reaching 28.76\%, and declines as adaptation spreads beyond the router. Under their shared budget, IA3 reaches 28.04\% while Houlsby reaches 25.39\%. The criterion transfers to Qwen1.5-MoE fine-tuned for mathematical reasoning, retaining 49.7\% mean accuracy over eleven benchmarks with half the experts removed. Structural pruning reduces memory by 49\% and per-token latency by 37\%. Lightweight router sensitivity therefore makes provably motivated, task-conditioned expert pruning practical at scale.
\end{abstract}

\section{Introduction}

MoE architectures route each token to a small subset of expert subnetworks, separating total parameter count from active computation \citep{shazeer2017outrageously,lepikhin2021gshard,fedus2022switch}. The saving is in computation rather than memory: every expert must remain resident even though only a few are active per token. Mixtral-8$\times$7B contains 46.7B parameters while activating 13B, and this gap restricts deployment on constrained hardware \citep{jiang2024mixtral}. Removing entire experts addresses the memory cost directly, but the central question is which experts can be removed.

Recent work locates a pruning signal in the router. \citet{chowdhury2024provably} prove that prioritizing experts with smaller router-norm changes during fine-tuning preserves test accuracy on a simplified MoE, and validate the criterion on vision models. The quantity is almost free to read once adaptation has occurred, but the original criterion assumes full fine-tuning. Fully fine-tuning a 46.7B-parameter MoE only to decide which experts to delete inverts the economics of pruning.

We ask whether the same signal can be recovered from an adaptation cheap enough to run at scale. A brief parameter-efficient adaptation (PEFT) acts as a measuring instrument: it places the model under task pressure, and the router records which experts that pressure recruits. We then discard the adapter and structurally prune the original checkpoint. Our contributions are:

\begin{itemize}
\item To our knowledge, no prior work combines router norm-change pruning with PEFT on MoE language models; router-only LoRA (0.002\% trainable) recovers the signal.
\item We show that the signal is governed by adapter location, size, and perturbation geometry. Confining LoRA to the router is best; larger adapters produce better rankings; under a shared budget, multiplicative IA3 outperforms additive Houlsby at maximal pruning.
\item We demonstrate transfer across benchmarks and to Qwen1.5-MoE under full mathematical fine-tuning. We also provide the first MMLU-Pro and BIG-Bench Hard evaluation of expert-pruned Mixtral, compare against recent one-shot methods on shared benchmarks, and measure near-linear deployment savings.
\end{itemize}

Training-free criteria score frozen checkpoints using routing frequency, activations, reconstruction, or expert weights \citep{lu2024notall,jaiswal2025finding,lasby2025reap,liu2026aimer}. Router sensitivity is complementary and task-conditioned: router-only adaptation trains 0.002\% of Mixtral's parameters, and deployments already using adaptation obtain the ranking as a byproduct. Appendices~\ref{sec:related} and~\ref{sec:compare} review this distinction and compare recent one-shot methods.

\section{Router sensitivity from lightweight adaptation}
\label{sec:mainmethod}

For an MoE layer with $E$ experts, the router has weight matrix $W_g\in\mathbb{R}^{E\times d}$. Its $e$-th row $w_e$ produces logit $\ell_e(x)=\langle w_e,x\rangle$, and top-$k$ routing selects the experts used for token representation $x$. Writing $w_e^{(0)}$ and $w_e^{(T)}$ for the router vectors before and after adaptation, we score expert $e$ by the magnitude of its router-norm revision,
\begin{equation}
s_e=\left|\,\lVert w_e^{(T)}\rVert_2-\lVert w_e^{(0)}\rVert_2\,\right|.
\label{eq:main-weight}
\end{equation}
We rank experts independently within each layer and remove those with the smallest scores. The absolute value treats both expansion and contraction as evidence that adaptation engaged an expert. The theorem covers full fine-tuning of a simplified MoE with signed norm changes; it motivates but does not guarantee our PEFT, absolute-change, or logit-delta extensions (Appendix~\ref{sec:scores}).

Methods that leave $W_g$ frozen still change the representations reaching it. For those methods, we collect router logits from the base and adapted models over $N$ held-out tokens and use
\begin{equation}
s_e^{\mathrm{logit}}=\frac{1}{\sqrt{N}}\left\lVert L^{\mathrm{adapt}}_{:,e}-L^{\mathrm{base}}_{:,e}\right\rVert_2.
\label{eq:main-logit}
\end{equation}
This places direct low-rank router updates, multiplicative IA3 modulation \citep{liu2022ia3}, and additive Houlsby adapters \citep{houlsby2019parameter} in one framework. LoRA rank $r$ indexes the adapter's internal width and parameter budget \citep{hu2022lora}. For the $8\times d$ Mixtral router, $r=16$ is a larger parameterization whose algebraic update rank remains capped at eight. We sweep $r\in\{2,4,8,16\}$ while scaling $\alpha$ with $r$.

Pruning is structural and one-shot. We delete selected experts and their router rows from the original pretrained checkpoint, retain top-$k$ routing over the survivors, and perform no retraining, distillation, or router calibration. The adapted model is only the instrument that produces the ranking; no adapter parameter survives into deployment.

\section{Experimental setting}

Our primary model is Mixtral-8$\times$7B-Instruct-v0.1, with 32 layers, eight experts per layer, and two active experts per token. Adaptation uses a generic text-generation instruction corpus at sequence length 1024. The controlled LoRA runs use 500 optimizer steps of 8-bit AdamW, learning rate $2\times10^{-4}$, weight decay 0.01, cosine decay, five warmup steps, bfloat16 compute, and a 4-bit frozen base. The reference attaches rank-8 LoRA only to router weights. We separately vary target modules and adapter size. IA3 and Houlsby share a separate matched 2,000-example, one-epoch budget; Appendix~\ref{sec:setup} gives their configurations.

We structurally remove one through four experts from every Mixtral layer and evaluate with LightEval \citep{fourrier2023lighteval}. MMLU-Pro is primary \citep{wang2024mmlupro}; GSM8K, ARC-Challenge, BIG-Bench Hard, and BoolQ test cross-benchmark behavior. We additionally study a 20-expert-per-layer Qwen1.5-MoE-A2.7B-Chat checkpoint \citep{qwen2024moe}, fully fine-tune it on mathematical reasoning, and evaluate pruning over eleven mathematics benchmarks. Full model, data, configuration, replication, and implementation details appear in the appendix.

\section{Results}
\label{sec:mainresults}

\begin{figure}[t]
\centering
\begin{minipage}[t]{0.49\linewidth}
\centering
\includegraphics[width=\linewidth]{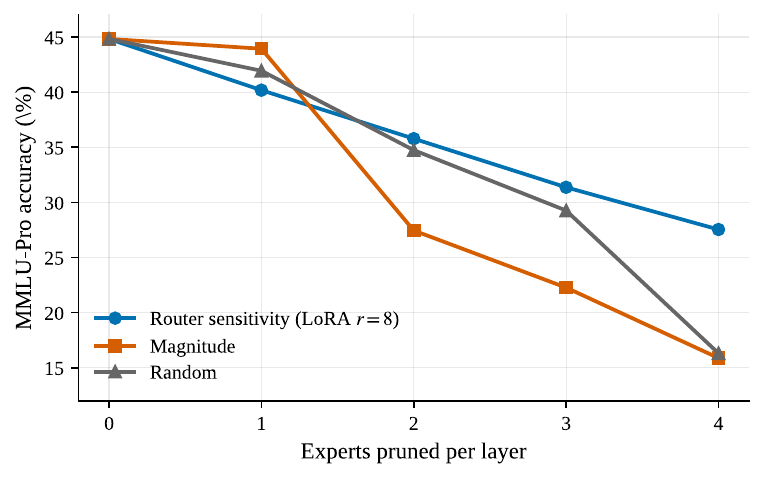}
\small (a) Selection criteria
\end{minipage}\hfill
\begin{minipage}[t]{0.49\linewidth}
\centering
\includegraphics[width=\linewidth]{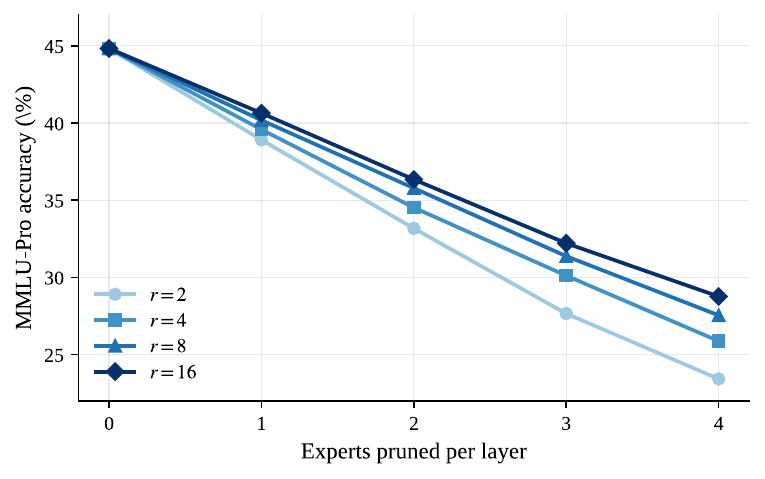}
\small (b) Router-only LoRA adapter size
\end{minipage}
\caption{MMLU-Pro accuracy after expert pruning on Mixtral-8$\times$7B-Instruct. (a) Router sensitivity decays quasi-linearly while magnitude and random selection collapse. (b) Larger adapter parameterizations produce better rankings at every pruning depth, with the endpoint ordering preserved across two independent runs.}
\label{fig:main-results}
\end{figure}

\paragraph{Router sensitivity survives lightweight adaptation.}
The unpruned model scores 44.83\% MMLU-Pro. Router-only LoRA at rank 8 retains 40.18\%, 35.78\%, 31.37\%, and 27.54\% as one through four experts are removed per layer. At one expert pruned, magnitude and random lead by 3.76 and 1.76 points; router sensitivity becomes advantageous beyond the marginal expert. At maximal compression, magnitude and random selection retain 15.88\% and 16.31\%. Router-guided pruning therefore retains roughly 1.7 times their accuracy, and its decay is quasi-linear across all five compression points, making the accuracy cost of a target compression ratio predictable.

\begin{table}[t]
\centering
\small
\caption{Controlled Mixtral results at 50\% expert pruning. The adapter is discarded after ranking, so every pruned model has the same 24.2B parameters.}
\label{tab:main-results}
\begin{tabular}{llcc}
\toprule
Selection & Score & Trainable & MMLU-Pro \\
\midrule
Magnitude & static weights & 0 & 15.88 \\
Random & none & 0 & 16.31 \\
\midrule
All-module LoRA, $r=8$ & weight delta & 0.26\% & 24.42 \\
Router LoRA, $r=2$ & weight delta & 0.0006\% & 23.42 \\
Router LoRA, $r=4$ & weight delta & 0.0011\% & 25.87 \\
Router LoRA, $r=8$ & weight delta & 0.0022\% & 27.54 \\
Router LoRA, $r=16$ & weight delta & 0.0045\% & \textbf{28.76} \\
IA3 & logit delta & $\sim$0.01\% & 28.04 \\
Houlsby & logit delta & $\sim$0.14\% & 25.39 \\
\bottomrule
\end{tabular}
\end{table}

\paragraph{The signal concentrates at the router.}
At matched rank, attaching LoRA to every linear module yields 24.42\% at four experts pruned, against 27.54\% when adaptation is confined to the router. Accuracy declines steadily as query and output projections, full attention, expert weights, and finally all linear modules are added. The optimizer otherwise gains ways to lower loss by changing what experts compute rather than which experts are recruited, diluting the measurement at the router.

\paragraph{Larger adapters recover a better ranking.}
Across router-only LoRA sizes $r=2,4,8,16$, every MMLU-Pro column is monotonically ordered at all four pruning depths. At maximal compression, accuracy rises from 23.42\% to 25.87\%, 27.54\%, and 28.76\%. Since the adapter is discarded, these models differ only in the experts retained. Larger adapter parameterizations therefore inject a richer task-conditioned revision and produce a more faithful ranking.
The endpoint ordering is replicated: rank 16 remains best and rank 2 worst in both runs at every pruning depth. Paired runs differ by 0.76 points on average and at most 2.24 points, so we treat 2.2 points as the resolution of a single comparison.

\paragraph{Constrained perturbations preserve the signal.}
IA3 reaches 28.04\% without writing to the router, within resolution of router-only LoRA. Under their shared budget, it leads Houlsby (25.39\%) by 2.65 points, above the 2.2-point resolution. Because this budget differs from LoRA's, the experiment does not isolate geometry. The ordering is consistent with constrained multiplicative modulation preserving router distinctions better than an unconstrained additive residual, suggesting separate roles for capacity and faithfulness.

\paragraph{The signal transfers.}
Across additional Mixtral benchmarks, ARC-Challenge and BIG-Bench Hard decline gently, GSM8K has a late cliff, and BoolQ is non-monotonic, exposing the adaptation corpus as a control surface for protected capabilities. On Qwen1.5-MoE fully fine-tuned for mathematics, router sensitivity retains 58.1\% and 49.7\% mean accuracy after five and ten of twenty experts are removed. Magnitude retains 55.2\% and 39.1\%; random pruning falls from 37.6\% to 4.4\%. The criterion thus transfers across model family, adaptation regime, and domain.

\begin{figure}[t]
\centering
\includegraphics[width=0.90\linewidth]{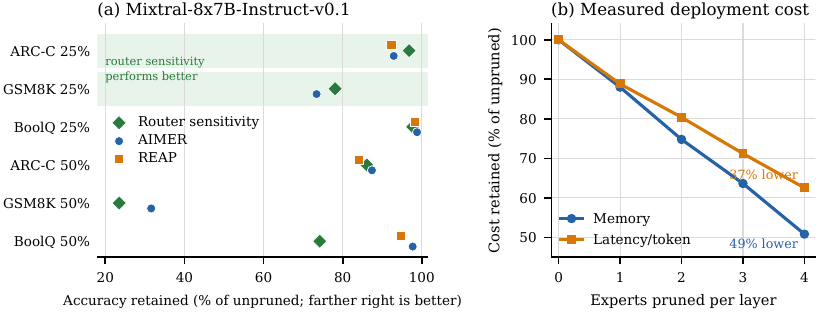}
\caption{Modern-method positioning and deployment cost. (a) Mixtral retention normalized to each study's unpruned baseline. Pale green: router sensitivity is better than AIMER and REAP. Cross-study values; unreported results omitted. (b) Memory and latency relative to unpruned Mixtral.}
\label{fig:main-compare-efficiency}
\end{figure}

\paragraph{Comparison with recent one-shot pruning methods.}
AIMER scores frozen-weight concentration, while REAP combines router weights with expert activation norms \citep{liu2026aimer,lasby2025reap}. All three use Mixtral-8$\times$7B-Instruct-v0.1, but harnesses differ, so Figure~\ref{fig:main-compare-efficiency}a normalizes to each study's unpruned baseline. At 25\% pruning, router sensitivity leads both on ARC-Challenge and AIMER on GSM8K, but trails AIMER by 1.1 points and REAP by 0.6 on BoolQ. At 50\%, it leads REAP by 2.1 points and trails AIMER by 1.3 on ARC-Challenge; both retain more on BoolQ, and AIMER retains more on GSM8K. Router sensitivity is therefore competitive at moderate compression while providing a task-conditioned signal from lightweight adaptation. Appendix~\ref{sec:compare} includes DiEP and MoE Pathfinder.

\section{Deployment efficiency}

Structural pruning yields measured deployment savings (Figure~\ref{fig:main-compare-efficiency}b). With 4-bit weights on one A100, pruning four experts per layer cuts memory from 24.2 to 12.3 GB (49\%) and latency from 40.3 to 25.2 ms (37\%), raising single-stream throughput by 60\%. Under this implementation, we attribute the latency gain to a smaller resident expert set and reduced memory traffic; top-2 routing and expert FLOPs remain unchanged, and the magnitude is implementation-dependent.

\section{Conclusion}

Router sensitivity turns lightweight adaptation into an expert-selection instrument: adapt on protected-capability data, rank by router change, discard the adapter, and prune the original checkpoint. Near-linear accuracy and deployment-cost curves make the tradeoff predictable. The monotone rank effect, geometry dependence, and sign question lie beyond the current guarantee. We release our fine-tuning and pruning code, training data, and evaluated checkpoints through the links in Appendix L.

\clearpage
\bibliographystyle{plainnat}
\bibliography{references}

\clearpage
\appendix
\input{appendix_full}

\end{document}

%% file: appendix_full.tex
\section{Related work}
\label{sec:related}

\textbf{Expert pruning in MoE models.} Because MoE layers hold most of a model's
parameters, removing whole experts is the most direct route to a smaller
checkpoint. Early work pruned experts progressively during task-specific
fine-tuning until a single expert remained per layer
\citep{chen2022taskspecific}, and \citet{koishekenov2023memory} showed that
language-specific experts in a multilingual translation model could be removed
without any further training. Subsequent criteria have grown more selective:
\citet{muzio2024seermoe} rank experts by accumulated router gate values,
\citet{lu2024notall} search for the expert subset that minimizes reconstruction
loss against the original layer output, and \citet{liu2024efficientexpert} use a
gradient-free evolutionary search over expert subsets. \citet{lasby2025reap}
combine gate value with expert activation norm and report near-lossless
compression at 50\% of experts removed on models up to 1T parameters.
\citet{jaiswal2025finding} survey sixteen criteria on Mixtral and find
activation-norm scoring strongest overall. \citet{liu2026aimer} instead score
the concentration pattern of frozen expert weights, requiring neither
calibration data nor activation statistics, and report balanced retention across
five MoE families including Mixtral. \citet{bai2025diep} relax the selection into a differentiable optimization that learns a nonuniform pruning rate per layer, and
\citet{yang2025pathfinder} cast expert retention as global path planning over activation trajectories. A related line prunes weights within experts rather than experts themselves
\citep{xie2024moepruner}. What unites these methods is that they score experts from a frozen checkpoint using calibration data, with no adaptation step.

\textbf{Router-derived importance signals.} Several of the above criteria read
the router, but they read its \emph{current state}: gate values on calibration
data \citep{muzio2024seermoe, lasby2025reap}, or the $\ell_2$ norm of each
expert's row in the gate matrix, which \citet{jaiswal2025finding} report to be
the strongest of the weight-based scores they evaluate. A different signal is
available when the model is adapted. \citet{chowdhury2024provably} define the
sensitivity of expert $e$ as the change in its router norm over fine-tuning,
$\Delta_e = \lVert w_e^{(T)} \rVert - \lVert w_e^{(0)} \rVert$, and prove that
removing the experts with smallest $\Delta_e$ preserves test accuracy. The
distinction matters for what the score can express: a static norm measures how
much the pretrained router relies on an expert, whereas $\Delta_e$ measures how
much adaptation to a task revises that reliance. Their analysis covers binary
classification on a simplified MoE, and their empirical validation uses vision
MoE models fine-tuned on image classification benchmarks. Adapting the router
alone is itself an established technique: \citet{hu2026whatgetsactivated} attach
LoRA to the gating projection of Mixtral, DeepSeek and Qwen to study expert
attribution, though they do not use the resulting weight changes to prune. We
take the sensitivity criterion as our starting point and ask what happens when
the adaptation producing $\Delta_e$ is made parameter-efficient rather than full,
and when the model is a MoE language model rather than a vision classifier.

\textbf{Expert merging and post-compression repair.} Rather than discarding
experts, merging collapses similar experts into one, clustering by router-logit
similarity and averaging weights \citep{li2023merge}. Merging has been reported
to beat pruning on multiple-choice benchmarks, but \citet{lasby2025reap} show it
introduces an irreducible error by tying gate values that the router previously
controlled independently, and that the advantage reverses on generative tasks.
Independently of the criterion used, some accuracy is usually recoverable
afterwards: \citet{xie2024moepruner} restore 99\% of baseline performance through
expert-wise distillation, and \citet{jaiswal2025finding} find iterative dropping
with interleaved retraining preferable to one-shot removal.
\citet{hyeon2026retrainingfree} argue more strongly that compression leaves the
router mismatched to the surviving experts, and recover much of the loss by
distilling into the router alone. They find this recovery far larger for
fine-grained architectures than for Mixtral, whose eight experts per layer admit
only $\binom{8}{2}$ routing configurations. We evaluate in the one-shot,
uncalibrated regime throughout, which isolates the quality of the selection
criterion from the quality of any subsequent repair, and
Appendix~\ref{sec:discussion} returns to what calibration would add.

\textbf{Parameter-efficient fine-tuning.} PEFT methods adapt a frozen model
through a small set of additional trainable parameters. LoRA \citep{hu2022lora}
adds a low-rank product to selected weight matrices, and QLoRA
\citep{dettmers2023qlora} makes this feasible on a single node by keeping the
base weights in 4-bit precision. Houlsby adapters
\citep{houlsby2019parameter} insert bottleneck MLPs with a residual connection,
prefix tuning \citep{li2021prefix} prepends learned vectors to each layer's
input, and IA3 \citep{liu2022ia3} learns per-channel multiplicative scalings of
inner activations. These differ in trainable parameter count, but for our
purposes the more consequential difference is geometric: LoRA modifies weights
additively in a low-rank subspace, Houlsby adapters modify activations additively
through a nonlinear bottleneck, and IA3 modifies them multiplicatively without
changing their direction. Appendix~\ref{sec:how} finds that this geometric distinction, and not the
parameter budget, is what orders the three as instruments.

\textbf{PEFT as an instrument rather than a training method.} The PEFT literature
evaluates these methods by how closely they approach full fine-tuning on task
loss, and existing guidance about where to attach adapters follows from that
objective. We use them differently. Here the adapted model is not the deliverable;
the fine-tuning run is an instrument, and what we read off it is which experts the
task recruits. The two objectives can favor different configurations, and
Appendix~\ref{sec:discussion} returns to the case where they visibly diverge.

\section{Measuring router sensitivity}
\label{sec:method}

\subsection{Preliminaries: MoE routing}
\label{sec:prelim}

We consider a decoder-only transformer in which the feed-forward block of each
layer is replaced by an MoE layer holding $E$ experts
$f_1, \dots, f_E$, each a map $\mathbb{R}^{d} \to \mathbb{R}^{d}$, together with a
router. The router is a single linear map with weight matrix
$W_g \in \mathbb{R}^{E \times d}$ whose $e$-th row $w_e \in \mathbb{R}^{d}$ is the
gate vector of expert $e$. For a token representation
$x \in \mathbb{R}^{d}$, the router produces logits
\begin{equation}
\ell(x) = W_g\, x \in \mathbb{R}^{E},
\qquad \ell_e(x) = \langle w_e, x \rangle ,
\end{equation}
and top-$k$ routing selects the index set $\mathcal{T}_k(x)$ of the $k$ largest
logits. Gate values are normalized over the selected experts alone, so the layer
output is
\begin{equation}
y(x) = \sum_{e \in \mathcal{T}_k(x)} g_e(x)\, f_e(x),
\qquad
g_e(x) = \frac{\exp \ell_e(x)}{\sum_{j \in \mathcal{T}_k(x)} \exp \ell_j(x)} .
\label{eq:moe}
\end{equation}
In Mixtral-8$\times$7B-Instruct, every one of the 32 layers carries $E = 8$
experts with $k = 2$ active per token. Because each expert is a full SwiGLU
feed-forward block while $w_e$ is a single $d$-dimensional vector, the router
accounts for a negligible share of parameters yet fully determines which experts
contribute to any given token.

\subsection{Sensitivity scores}
\label{sec:scores}

\textbf{Router norm change.} The criterion we build on measures how far
adaptation moves an expert's gate vector. Writing $w_e^{(0)}$ for the gate vector
of expert $e$ in the pretrained model and $w_e^{(T)}$ for its value after
fine-tuning, \citet{chowdhury2024provably} define the signed change
\begin{equation}
\Delta_e = \bigl\lVert w_e^{(T)} \bigr\rVert_2 - \bigl\lVert w_e^{(0)} \bigr\rVert_2
\label{eq:signed}
\end{equation}
and retain the experts with the largest values of $\Delta_e$. The intuition is
that fine-tuning revises the router's reliance on experts the task actually
recruits, while experts the task does not use are left near their pretrained
gating.

We score experts by the \emph{magnitude} of this revision,
\begin{equation}
s_e = \bigl\lvert\, \bigl\lVert w_e^{(T)} \bigr\rVert_2 - \bigl\lVert w_e^{(0)} \bigr\rVert_2 \,\bigr\rvert ,
\label{eq:abs}
\end{equation}
and prune the experts with the smallest $s_e$. The two scores agree whenever
gate norms only grow, and diverge when a norm contracts: such an expert has a
large $s_e$ and is retained under Equation~\ref{eq:abs}, but has a strongly
negative $\Delta_e$ and would be an early pruning candidate under
Equation~\ref{eq:signed}. We adopt the direction-agnostic form because our
question is how much adaptation perturbs the router's treatment of an expert, not
in which direction, and because a contracting gate is still evidence that the
task engaged that expert. We note that the theoretical guarantee is stated for
the signed quantity, and return in Appendix~\ref{sec:discussion} to whether the
sign carries usable information.

\textbf{Logit deltas when the router is frozen.} Equations~\ref{eq:signed} and
\ref{eq:abs} require $W_g$ to be trainable. Several adaptation methods leave the
router weights untouched and alter only the representations reaching it, in
which case $s_e = 0$ for every expert and the score is vacuous. The routing
behavior nonetheless changes, because the logits in Equation~\ref{eq:moe} are a
function of the incoming representation as much as of $W_g$. We therefore extend
the criterion to act on logits rather than weights. Let
$\mathcal{X}$ be a held-out batch of $N$ token representations, and let
$L^{\mathrm{base}}, L^{\mathrm{adapt}} \in \mathbb{R}^{N \times E}$ collect the
router logits produced by the pretrained and adapted models on $\mathcal{X}$.
With $\Delta L = L^{\mathrm{adapt}} - L^{\mathrm{base}}$, we define
\begin{equation}
s_e^{\mathrm{logit}} = \frac{1}{\sqrt{N}} \bigl\lVert \Delta L_{:,e} \bigr\rVert_2 ,
\label{eq:logit}
\end{equation}
the root-mean-square shift in expert $e$'s logit across tokens. The
normalization makes the score comparable across batch sizes. Equation~\ref{eq:logit}
reduces to a measure of the same underlying quantity as
Equation~\ref{eq:abs}, the extent to which adaptation revises the router's
treatment of an expert, but reads it downstream of the gate rather than in its
weights, and is therefore defined for every method we consider.

\subsection{Adaptation families and their perturbation geometry}
\label{sec:families}

Throughout, a parameter-efficient method is applied for a short fine-tuning run
whose only purpose is to induce a measurable change in routing. We consider three
families, which differ less in parameter count than in
how they perturb the quantities entering Equation~\ref{eq:moe}.

\textbf{LoRA} \citep{hu2022lora} reparameterizes a weight matrix $W$ as
$W + BA$ with $B \in \mathbb{R}^{d_{\mathrm{out}} \times r}$,
$A \in \mathbb{R}^{r \times d_{\mathrm{in}}}$ and $r \ll \min(d_{\mathrm{in}}, d_{\mathrm{out}})$.
Applied to $W_g$, it perturbs the gate vectors directly and additively within an
$r$-dimensional subspace, so $w_e$ changes and Equation~\ref{eq:abs} applies.
Applied elsewhere, it perturbs the representation $x$ that reaches an unchanged
router. The rank $r$ sets the adapter's internal width and parameter budget and
is swept in Appendix~\ref{sec:rank}.

\textbf{IA3} \citep{liu2022ia3} learns a per-channel scaling vector $\alpha$ and
replaces $x$ by $x \odot \alpha$. Router logits become
$\ell(x) = W_g (x \odot \alpha)$, so gating is modulated multiplicatively while
$W_g$ is untouched. Because scaling acts coordinate-wise on the input, the
induced logit shift is a reweighting of contributions already present in
$\langle w_e, x \rangle$.

\textbf{Houlsby adapters} \citep{houlsby2019parameter} insert a bottleneck
$x \mapsto x + U\,\sigma\!\left(D\,\mathrm{LN}(x)\right)$ with
$D \in \mathbb{R}^{r \times d}$, $U \in \mathbb{R}^{d \times r}$ and a
nonlinearity $\sigma$. The perturbation is additive in representation space and
passes through a nonlinearity, so it is unconstrained in direction: the residual
may point anywhere in $\mathbb{R}^{d}$, including directions with no relation to
any $w_e$.

Only the first of these updates the router weights, so LoRA on the gate admits
either score while IA3 and Houlsby require Equation~\ref{eq:logit}. The three
perturbation geometries, low-rank additive in weight space, multiplicative in
representation space, and unconstrained additive in representation space, are
compared in Appendix~\ref{sec:how}, where the results are consistent with a role
for geometry beyond parameter budget without isolating it as the sole cause.

\subsection{One-shot pruning procedure}
\label{sec:prune}

\begin{figure}[t]
\centering
\includegraphics[width=0.86\linewidth]{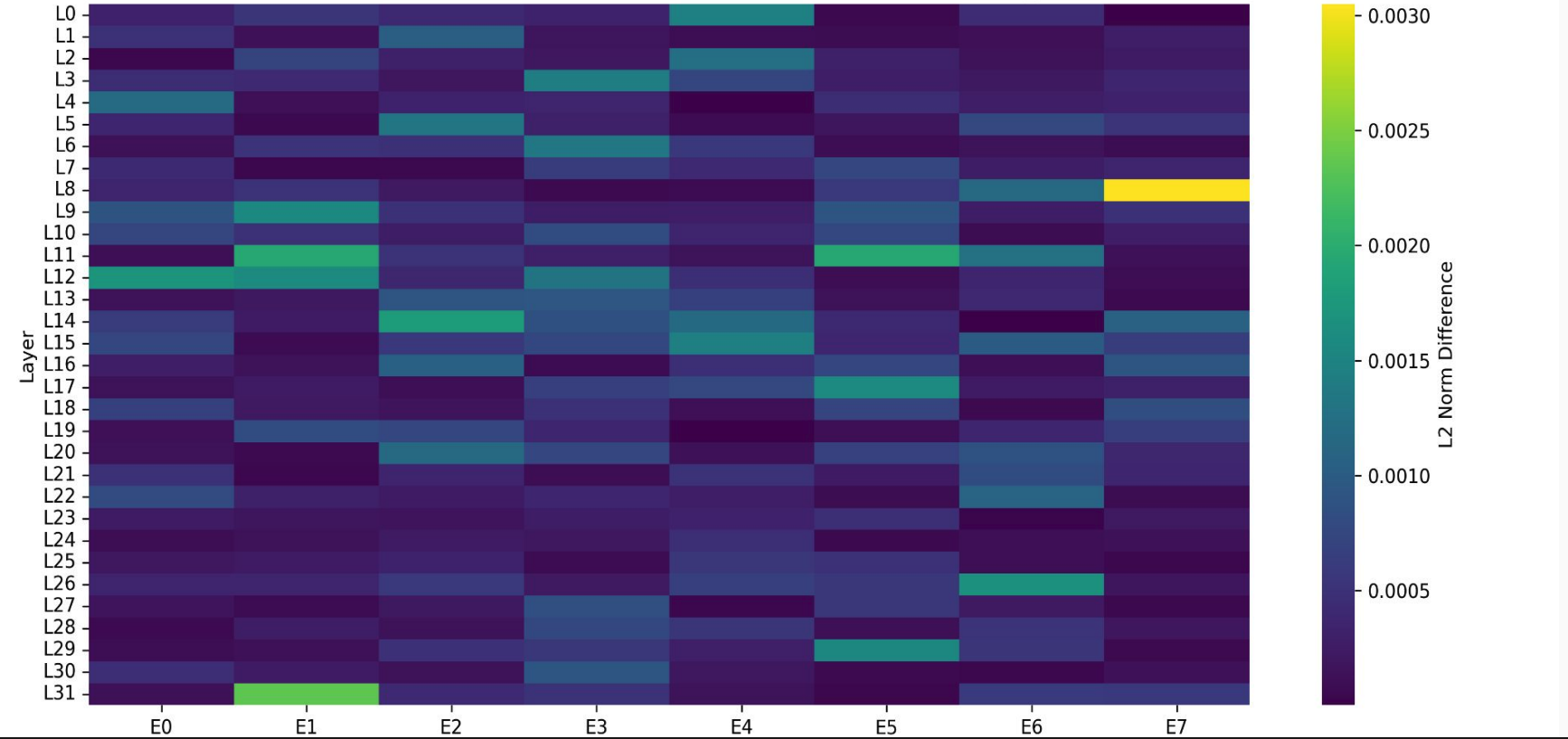}
\caption{Router sensitivity across Mixtral-8$\times$7B-Instruct: the per-expert
$\ell_2$ norm change $s_e$ of Equation~\ref{eq:abs}, for all 32 layers and 8
experts, from an all-module LoRA adaptation at rank~16. The other adaptation
runs in this study produce maps with the same scattered structure, differing
mainly in overall scale. Sensitivity is not concentrated: most cells sit near
their layer's median while a few expert-layer pairs stand out by an order of
magnitude (the strongest here is expert~7 of layer~8), and no single expert
column or band of layers dominates. This scatter is why experts are ranked
\emph{within} each layer rather than against a global threshold
(Appendix~\ref{sec:prune}).}
\label{fig:heatmap}
\end{figure}

Scores are computed and applied independently per layer (Figure~\ref{fig:heatmap}). Within layer $\ell$ we
rank its $E$ experts by $s_e$ (or $s_e^{\mathrm{logit}}$), remove the $n$ experts
with the smallest scores, and delete their parameters from the checkpoint. We
prune the same number $n$ from every layer, so a model with $n$ pruned retains
$E - n$ experts throughout. The corresponding rows of $W_g$ are removed with
them, which leaves Equation~\ref{eq:moe} well-formed on the surviving experts:
top-$k$ selection and gate normalization are simply carried out over $E - n$
logits instead of $E$. For Mixtral with $k = 2$, pruning up to $n = 4$ leaves at
least two experts per layer and requires no change to the routing rule.

Two properties of this procedure are worth stating explicitly. Pruning is applied
to the original pretrained checkpoint, not to the adapted one: the adapter supplies
the ranking and is then discarded, so the deployed model contains no adapter
parameters. And pruning is one-shot, with no retraining, distillation, or router
recalibration afterwards. This isolates the quality of the selection criterion,
which is what we set out to measure, at the cost of forgoing recoverable
accuracy; Appendix~\ref{sec:discussion} discusses what a repair step would add.

\section{Experimental setup}
\label{sec:setup}

\textbf{Models.} Our primary model is Mixtral-8$\times$7B-Instruct-v0.1
\citep{jiang2024mixtral}, a decoder-only transformer with 32 layers, $E = 8$
experts per layer and $k = 2$ active per token, totalling 46.7B parameters of
which 13B are active. It is a deliberate choice for this study: with only eight
coarse experts per layer, each removal is a large intervention, so a pruning
criterion has little room to be accidentally right. To test whether the criterion
survives a change of model family, training regime and domain, we additionally
use a 20-expert-per-layer Qwen1.5-MoE-A2.7B-Chat checkpoint
\citep{qwen2024moe} with $k = 4$ active, as a fine-grained counterpart to Mixtral.

\textbf{Data.} All Mixtral adaptation runs use
an anonymized augmented supervised
fine-tuning corpus of general text-generation instructions, training on the
completion field at a maximum sequence length of 1024 tokens. The corpus is
deliberately generic: we ask which experts a broad instruction distribution
recruits, rather than tailoring the signal to any evaluation benchmark. The
mathematical-reasoning runs of Appendix~\ref{sec:generalize} instead use a mixture
of MathInstruct and Belle Math, making that transfer test a change of domain as
well as of model.

\textbf{Adaptation configurations.} Except where a factor is deliberately varied,
runs share one recipe: 500 optimizer steps of 8-bit AdamW at learning rate
$2\times10^{-4}$, weight decay 0.01, a cosine schedule with 5 warmup steps,
bfloat16 compute, per-device batch size 1 with gradient accumulation, and LoRA
dropout 0.1, from seed 3407. Accumulation is 20 for the router-only reference and
is held fixed within each sweep; per-run values are given in
Appendix~\ref{app:configs}. Mixtral adaptation runs hold the base weights in
4-bit precision following QLoRA \citep{dettmers2023qlora}; the Qwen router
adaptation runs unquantized. Our reference configuration applies LoRA of rank
$r = 8$ with $\alpha = 8$ to the router weights alone. From that reference we vary
one factor at a time: rank over $\{2, 4, 8, 16\}$ with the target set fixed
(Appendix~\ref{sec:rank}), and the target module set over router-only, router with
query and output projections, router with full attention, router with expert
feed-forward weights, and all linear modules, with rank fixed at 8
(Appendix~\ref{sec:where}). Every configuration in this sweep includes the
router, since excluding it leaves $W_g$ untouched and Equation~\ref{eq:abs}
identically zero. Such configurations could in principle be scored with
Equation~\ref{eq:logit} instead, and we return to this gap in
Appendix~\ref{sec:discussion}.

The multiplicative and additive-adapter runs use a fixed budget of 2000 training
examples for one epoch at learning rate $10^{-4}$, with context length 2048,
micro-batch size 1 and gradient accumulation 16. IA3 \citep{liu2022ia3} attaches
scaling vectors to the router, the attention projections and the expert
feed-forward weights; Houlsby adapters \citep{houlsby2019parameter} use a
reduction factor of 16 with a GELU nonlinearity. Because this budget differs from
that of the LoRA sweeps, Appendix~\ref{sec:how} compares adaptation geometries
under their own settings rather than under a matched budget, and we read those
results accordingly.

\textbf{Full fine-tuning reference.} One setting updates every parameter rather
than an adapter. On Qwen we fully fine-tune the 20-expert model on mathematical
reasoning, which is feasible only with aggressive offloading: DeepSpeed ZeRO
stage 3 with both optimizer states and parameters offloaded to CPU, gradient
checkpointing, and an effective batch size of 512 formed from micro-batches of 1
with 256 accumulation steps, using AdamW at learning rate $2\times10^{-4}$ with
weight decay 0.01. Sensitivity is then read from Equation~\ref{eq:abs} directly,
since $W_g$ is trainable.

\textbf{Pruning implementation.} Pruning is structural rather than masked. For
each layer we delete the selected experts from the module list and remove the
corresponding rows of $W_g$, so the router's output dimension shrinks from $E$ to
$E-n$ and top-$k$ selection with softmax renormalization proceeds over the
surviving experts exactly as in Equation~\ref{eq:moe}. We remove the same number
$n \in \{1,2,3,4\}$ from all 32 layers, corresponding to 12.5\% through 50\% of
experts. Pruning is applied to the original pretrained checkpoint rather than to
the adapted one, so no adapter parameters survive into any evaluated model.
Pruned models are saved with the foundation model's tokenizer and a corrected
expert count in the configuration, then evaluated as ordinary checkpoints.

\textbf{Evaluation.} We evaluate with LightEval \citep{fourrier2023lighteval}.
MMLU-Pro \citep{wang2024mmlupro} is the primary benchmark throughout, chosen
because its ten-option format and reasoning-heavy questions leave less headroom
for a degraded model to guess well than four-option multiple choice does; the
unpruned Mixtral baseline is 44.83\%. Appendix~\ref{sec:generalize} widens the
evaluation to GSM8K \citep{cobbe2021gsm8k}, ARC-Challenge \citep{clark2018arc},
BIG-Bench Hard \citep{suzgun2023bbh} and BoolQ \citep{clark2019boolq}, which span
chain-of-thought arithmetic, science question answering, multi-step reasoning and
binary reading comprehension, and which we use to test whether the shape of the
accuracy curve is a property of the criterion or of MMLU-Pro. The Qwen study is
evaluated on eleven mathematical-reasoning benchmarks covering college, high
school and elementary mathematics, algebra, precalculus, probability, induction
and MathQA.

\textbf{Implementation and hardware.} Adaptation uses PEFT and TRL's
\texttt{SFTTrainer} over HuggingFace Transformers, with bitsandbytes supplying
4-bit base weights and the 8-bit optimizer. Experiments ran on NVIDIA A100-SXM4
GPUs with 80\,GB HBM2 under CUDA 12.8 and driver 550.54.15, using a single GPU
for adaptation and up to three for pruning and serialization of the full
checkpoint. The memory and latency measurements of Appendix~\ref{sec:deploy} are
taken with Optimum-Benchmark \citep{moutawwakil2023optimum} under 4-bit weights
on a single A100, which is why the unpruned model is reported at 24.2\,GB rather
than at its half-precision footprint. Pruned
checkpoints, run configurations, and the scoring and pruning code are released.

\textbf{Measurement resolution.} Three configurations were run twice over the
course of the project under nominally identical settings. Across the twelve
resulting pairs of accuracies, repeated runs differ by 0.76 points on average and
by at most 2.24. We take 2.2 points as the resolution of a single comparison in
what follows, and distinguish throughout between differences that exceed it and
orderings that are supported instead by their consistency across many
comparisons. Appendix~\ref{sec:robustness} reports these replications in full.

\textbf{Two measurement pipelines.} Results were produced in two phases whose
evaluation settings differ. The first, from which every result in
Sections~\ref{sec:where} through \ref{sec:how} is drawn, uses the recipe above
and gives an unpruned MMLU-Pro baseline of 44.83\%. The second, reported in
Appendix~\ref{sec:robustness} and in the benchmark study of
Appendix~\ref{sec:generalize}, comprises four independently configured runs that
differ in rank, target modules, step budget, batch shape, weight decay and
schedule, and that additionally use gradient clipping at maximum norm 0.3,
sequence packing and length-grouped batching; their unpruned baselines are
45.35\% and 45.57\%. The Qwen study is a third setting with its own model,
training regime and benchmarks, and is reported against its own unpruned scores
throughout. Every comparison is computed within a setting against that setting's
own baseline, and we do not mix settings in any single table or figure. Full
configurations appear in Appendix~\ref{app:configs}.

\section{Where does the pruning signal live?}
\label{sec:where}

\begin{figure}[t]
\centering
\includegraphics[width=0.8\linewidth]{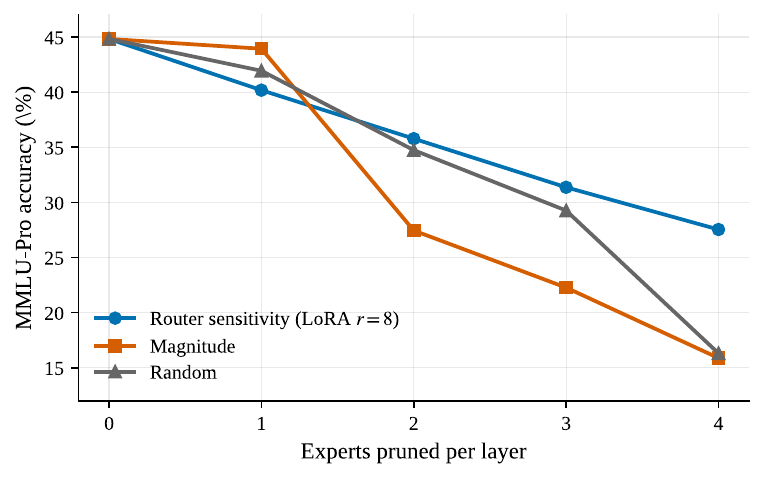}
\caption{MMLU-Pro accuracy as experts are removed from every layer of
Mixtral-8$\times$7B-Instruct, by selection criterion. Router sensitivity is the
router-only LoRA reference at rank~8 (Table~\ref{tab:where}); all curves share
the 44.83\% unpruned baseline at zero experts pruned. Router-guided selection
decays quasi-linearly, whereas magnitude and random selection collapse once more
than one expert per layer is removed.}
\label{fig:decay}
\end{figure}

The criterion of Appendix~\ref{sec:scores} says to read expert importance off the
router, but it does not say how the router should be made to move. This section
fixes the adapter rank at 8 and varies only which modules it is attached to, so
the adapted rows of Table~\ref{tab:where} differ in nothing except the location
of the adaptation. Two selection rules that require no adaptation at all are
included as a floor. Figure~\ref{fig:decay} plots the resulting accuracy against pruning depth.

\begin{table}[t]
\centering
\small
\caption{MMLU-Pro accuracy (\%) after removing $n$ experts from every layer of
Mixtral-8$\times$7B-Instruct. The unpruned model scores 44.83\% at 46.7B
parameters. All adapted rows use LoRA of rank 8 and include the router; rows
below the first add further target modules. Remaining parameter counts are
identical across selection criteria at a given $n$. Bold marks the best adapted
configuration in each column.}
\label{tab:where}
\begin{tabular}{lcccc}
\toprule
& \multicolumn{4}{c}{Experts pruned per layer} \\
\cmidrule(lr){2-5}
Selection criterion & 1 & 2 & 3 & 4 \\
& (41.1B) & (35.4B) & (29.8B) & (24.2B) \\
\midrule
\multicolumn{5}{l}{\emph{Without adaptation}} \\
Magnitude & 43.94 & 27.45 & 22.27 & 15.88 \\
Random & 41.94 & 34.74 & 29.26 & 16.31 \\
\midrule
\multicolumn{5}{l}{\emph{Router sensitivity, LoRA $r=8$}} \\
Router only & 40.18 & \textbf{35.78} & \textbf{31.37} & \textbf{27.54} \\
\quad + $q$, $o$ & 40.06 & 35.32 & 30.81 & 26.74 \\
\quad + $q$, $k$, $v$, $o$ & 39.94 & 34.98 & 30.40 & 25.93 \\
\quad + expert MLP & 39.82 & 34.85 & 29.97 & 24.97 \\
\quad + all linear modules & 39.87 & 34.86 & 30.04 & 24.42 \\
\bottomrule
\end{tabular}
\end{table}

\textbf{Confining adaptation to the router improves the signal.}
At matched rank, data, and training schedule, attaching LoRA to every linear module yields 24.42\% MMLU-Pro accuracy with four experts pruned per layer, compared with 27.54\% when LoRA is confined to the router.
The gap is 3.1 points at equal rank, equal data, and an identical pruned model
size; the two runs differ only in where the adapter was attached. Adapting the
whole network is the stronger configuration for producing a capable fine-tuned
model, and is what the parameter-efficient fine-tuning literature recommends when
task loss is the objective. It is the weaker configuration for deciding what to
delete.

\textbf{The penalty grows with the modules adapted and with pruning depth.} At
four experts pruned the five adapted rows form a ladder: 27.54\% for the router
alone, then 26.74\%, 25.93\% and 24.97\% as query and output projections, full
attention, and the expert feed-forward weights are added, and 24.42\% with every
linear module adapted. At shallower pruning the ordering is not strict. The five
configurations span 0.36 points at one expert pruned, 0.93 at two and 1.40 at
three, and within those bands the two largest-footprint configurations exchange
places, and no single step of the ladder exceeds the measurement resolution. What
the data support is the span rather than the individual rungs: the 3.1 points
between confining the adapter to the router and spreading it across the whole
network is the one comparison here larger than repeated-run variation. Target
module choice is close to irrelevant when one expert is removed and worth three
points when four are, because a criterion only needs to be right about the
marginal expert until it has to be right about half of them.

\textbf{Reference points without adaptation.} Neither adaptation-free rule
survives past the first removal. Magnitude selection is the best entry in the
$n=1$ column at 43.94\%, since every layer holds one expert the pretrained model
already weights lightly and any criterion correlated with that weighting will
find it, but it falls 16.5 points at $n=2$ against 4.4 for router sensitivity and
ends at 15.88\%. Random selection ends at 16.31\% and, in the middle of the
range, is the better of the two, which indicates that small weight magnitude is
not merely a weak indicator of redundancy but an actively misleading one. We do
not treat these rules as competitive baselines; they establish that the
comparisons in the rest of the paper are made above a floor rather than against
one.

\textbf{Interpretation.} The router is the point at which revised expert recruitment becomes observable, whether the change is induced by modifying the router weights directly or by changing the representations reaching them. When the adapter also reaches attention and expert weights,
the same reduction in loss becomes available by changing what the experts compute
rather than which are consulted. The router then moves less, and less
distinctively, and the measurement is diluted by the alternatives available to
the optimizer. This reading predicts the two results that follow: capacity spent
on the router should sharpen the signal (Appendix~\ref{sec:rank}), and adaptations
that reach the router only indirectly should be judged by how faithfully they
preserve its structure rather than by how many parameters they train
(Appendix~\ref{sec:how}).

\section{How much adaptation capacity is needed?}
\label{sec:rank}

\begin{figure}[t]
\centering
\includegraphics[width=0.8\linewidth]{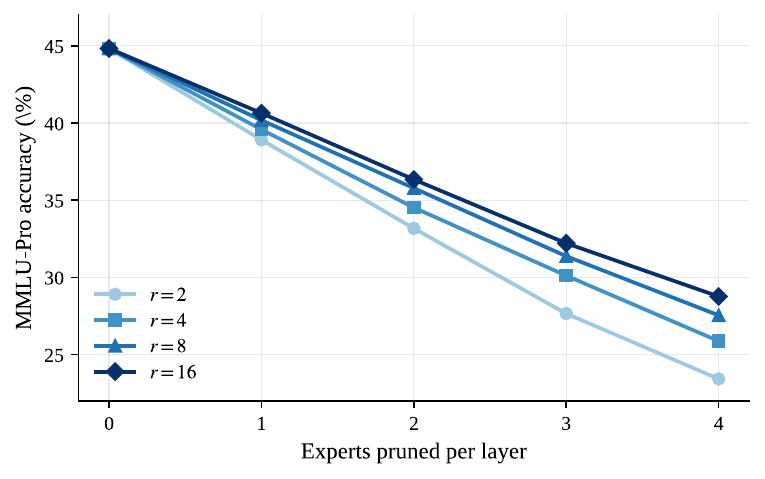}
\caption{MMLU-Pro accuracy after pruning for router-only LoRA at ranks
$r\in\{2,4,8,16\}$ (Table~\ref{tab:rank}). Each curve lies above the next-lower
rank at all four pruning depths, and the spread widens with pruning depth. The
adapter is discarded after ranking, so the curves describe models of identical
size that differ only in which experts were kept.}
\label{fig:rank}
\end{figure}

Appendix~\ref{sec:where} fixed the adapter rank and moved its location. This
section does the reverse: the adapter stays on the router and we sweep its rank
over $\{2, 4, 8, 16\}$, scaling $\alpha$ with $r$ and holding data, schedule and
step budget constant, in order to isolate the effect of adapter size (Figure~\ref{fig:rank}). Rank
sets the internal width and parameter budget of the adapter, so it controls how
rich a task-induced revision the adapter can inject into the router. Since the
Mixtral router has only eight rows, the algebraic rank of its update is capped
at eight; $r=16$ denotes a larger parameterization and a different optimization
path, not an update matrix with rank greater than eight.

\begin{table}[t]
\centering
\small
\caption{MMLU-Pro accuracy (\%) after pruning, for router-only LoRA at four
ranks. Trainable parameters are given as a percentage of the 46.7B-parameter
model. The unpruned model scores 44.83\%.}
\label{tab:rank}
\begin{tabular}{lccccc}
\toprule
& & \multicolumn{4}{c}{Experts pruned per layer} \\
\cmidrule(lr){3-6}
Rank & Trainable & 1 & 2 & 3 & 4 \\
\midrule
$r=2$ & 0.0006\% & 38.91 & 33.17 & 27.65 & 23.42 \\
$r=4$ & 0.0011\% & 39.58 & 34.52 & 30.11 & 25.87 \\
$r=8$ & 0.0022\% & 40.18 & 35.78 & 31.37 & 27.54 \\
$r=16$ & 0.0045\% & \textbf{40.64} & \textbf{36.34} & \textbf{32.21} & \textbf{28.76} \\
\bottomrule
\end{tabular}
\end{table}

\textbf{Accuracy after pruning increases monotonically with rank.} Every column
of Table~\ref{tab:rank} is ordered, without exception: rank 4 beats rank 2, rank
8 beats rank 4 and rank 16 beats rank 8 at all four pruning depths. No single
doubling clears the measurement resolution on its own; what supports the trend is
that all twelve adjacent comparisons point the same way, and that the endpoints
are separated by 5.34 points at four experts pruned. This is worth stating
carefully, because rank here does not act on the pruned model. The
adapter is discarded once experts have been ranked, and the four rows of
Table~\ref{tab:rank} describe models with identical architecture and identical
parameter counts, differing only in which experts were removed. Rank changes the
selection, not the artifact. A monotone trend therefore says that larger
adapters identify better pruning candidates, not that they leave behind a better
model.

\textbf{Returns diminish with each doubling.} Averaged across pruning depths,
moving from rank 2 to rank 4 is worth 1.73 points, rank 4 to rank 8 a further
1.20, and rank 8 to rank 16 a further 0.77. At the most aggressive setting the
same pattern holds with larger absolute values: $+2.45$, $+1.67$ and $+1.22$
points as rank doubles. The curve is concave in $\log r$, and the practical
reading is that rank 8 captures most of what is available, with rank 16 buying a
final point at twice the adapter size. Since even rank 16 trains 0.0045\% of the
model, the argument for stopping at 8 is not cost but the shape of the curve.

\textbf{Capacity matters more the more is pruned.} The spread between the
smallest and largest rank widens with pruning depth: 1.73 points at one expert
pruned, 3.17 at two, 4.56 at three and 5.34 at four. This mirrors the widening
observed across target modules in Appendix~\ref{sec:where}, and for the same
reason. Ranking the single least-used expert is a coarse decision that a
small adapter can get right; ordering half the experts in a layer benefits from
the richer task-conditioned revision injected by a larger parameterization. Rank-2 selection is
in fact worse than random removal at one, two and three experts pruned, and only
overtakes it at four, where random selection collapses. An adapter too small to
express the revision the task induces is not a weak instrument but a misleading
one.

\textbf{Interpretation.} Rank controls the adapter's internal width and parameter
budget. A smaller adapter injects a narrower task-conditioned revision, while a
larger parameterization can register more of the distinctions the task exposes
between experts. The monotone trend indicates that no adapter size we tested
overfits the selection: within this range, a larger adapter yields a more faithful
ranking. Where that trend would turn is not determined by these experiments.
Appendix~\ref{sec:how} turns instead to adaptations that cannot write to $W_g$ at
all.

\section{How should adaptation perturb the model?}
\label{sec:how}

\begin{figure}[t]
\centering
\includegraphics[width=0.8\linewidth]{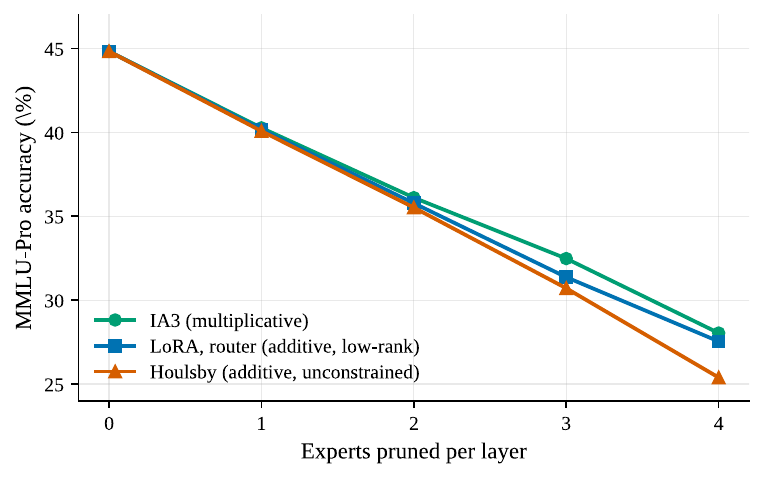}
\caption{MMLU-Pro accuracy after pruning by adaptation form
(Table~\ref{tab:how}). IA3, which never writes to the router, tracks router-only
LoRA within the measurement resolution at every depth, while the unconstrained
additive Houlsby adapter separates downward as pruning deepens under its tested budget.}
\label{fig:geometry}
\end{figure}

The two preceding sections varied where an adapter is attached and how much
capacity it has, holding its form fixed. This section varies the form (Figure~\ref{fig:geometry}). LoRA
writes a low-rank additive update into $W_g$ itself. IA3 leaves $W_g$ untouched
and rescales the representation reaching it, coordinate by coordinate. Houlsby
adapters leave $W_g$ untouched and add a nonlinear residual to that
representation. Only the first admits the weight score of
Equation~\ref{eq:abs}; the other two are read through the logit score of
Equation~\ref{eq:logit}. These runs use the budget given in
Appendix~\ref{sec:setup}, which differs from that of the LoRA sweeps, so
Table~\ref{tab:how} compares three geometries each under its own settings rather
than under a matched budget.

\begin{table}[t]
\centering
\small
\caption{MMLU-Pro accuracy (\%) after pruning, by adaptation form. The
unpruned model scores 44.83\%. LoRA is the router-only reference at rank 8 and is
scored on router weights; IA3 and Houlsby leave $W_g$ frozen and are scored on
router logits. IA3 and LoRA differ by less than the measurement resolution at
every depth; Houlsby trails both under the tested budgets.}
\label{tab:how}
\begin{tabular}{llcccc}
\toprule
& & \multicolumn{4}{c}{Experts pruned per layer} \\
\cmidrule(lr){3-6}
Adaptation & Perturbation & 1 & 2 & 3 & 4 \\
\midrule
IA3 & multiplicative, diagonal & 40.27 & 36.11 & 32.48 & 28.04 \\
LoRA (router) & additive, low-rank & 40.18 & 35.78 & 31.37 & 27.54 \\
Houlsby & additive, unconstrained & 40.08 & 35.52 & 30.71 & 25.39 \\
\bottomrule
\end{tabular}
\end{table}

\textbf{An adaptation that never writes to the router matches one that does.}
IA3 leads at every pruning depth, by 0.09, 0.33, 1.11 and 0.50 points over
router-only LoRA. None of these margins clears the measurement resolution of
Appendix~\ref{sec:setup}, and the budgets are unmatched, so we read the table as a
tie rather than a win. The tie is itself the finding: a method that modifies no
router parameter identifies prunable experts as well as one that modifies them
directly. The criterion does not require writing to the object being measured. It
requires only that the object be observed while the model is under task
pressure.

\textbf{The constrained perturbation preserves a cleaner signal.} Houlsby
adapters trail at every depth and the gap widens with pruning, from 0.10 points
at one expert removed to 2.15 at four. The endpoint gap to IA3, 2.65 points, is
the one comparison in this section that exceeds the measurement resolution, and
the widening is consistent across all four depths. Because the configurations
are unmatched, this ordering does not isolate geometry as the cause. Measured end to end, accuracy
falls 14.69 points between one and four experts pruned under Houlsby, against
12.64 for LoRA and 12.23 for IA3. Houlsby is nonetheless far above the
adaptation-free floor of Appendix~\ref{sec:where}, so its selections carry real
information; they are simply noisier.

\textbf{The scoring rule alone does not explain the ordering.}
IA3 and Houlsby are both scored using logit deltas on held-out data, whereas
LoRA is scored using router-weight changes. If the logit score alone were
responsible for IA3's performance, Houlsby would be expected to show a similar
advantage over LoRA. Instead, Houlsby performs worst at every pruning depth
despite using the same scoring rule as IA3. This suggests that differences in
how the adaptation methods perturb the model contribute to the observed
ordering. Because the methods are trained under unmatched budgets and
configurations, however, the experiment does not isolate perturbation geometry
as the sole causal factor.

\textbf{Reconciling this with Appendix~\ref{sec:where}.} IA3 attaches to the
router, the attention projections and the expert feed-forward weights, which is
the spread configuration that Appendix~\ref{sec:where} found harmful for LoRA. The
two results are consistent once dilution is understood as a function of capacity
available away from the router rather than of mere presence there. LoRA at rank 8
on all modules gives attention and experts substantial capacity, and the
optimizer can lower loss by changing what experts compute instead of which are
consulted, leaving the router less distinctly moved. IA3 assigns each module a
single scaling vector, on the order of 0.01\% of parameters in total, which is
too little capacity to absorb the task anywhere. Houlsby sits between the two in
budget yet performs worst, which shows that capacity alone does not order these
methods either.

\textbf{Interpretation: constraint as an instrument property.} The three
geometries differ in how freely they can reorganize routing for reasons unrelated
to expert utility. IA3 multiplies the representation by a vector shared across
every expert in a layer, so it cannot introduce a distinction between two experts
that their existing weight vectors do not already express; it can only amplify or
attenuate distinctions that are there. LoRA writes directly into $W_g$ within an
$r$-dimensional subspace, which is more free but still bounded. A Houlsby
residual passes through a nonlinearity and may point anywhere in representation
space, including directions that happen to align with one expert's gate vector
for reasons having nothing to do with the task. Read alongside
Appendix~\ref{sec:rank}, this suggests that capacity and faithfulness are separate
axes. Within a fixed geometry, more capacity gives a better ranking, since the
adapter can express more of the revision the task actually induces. Across
geometries, less freedom gives a better ranking, since the perturbation is more
tightly bound to structure the model already had. An instrument should be
expressive enough to register the effect and constrained enough not to create
one.

\section{Does the signal generalize?}
\label{sec:generalize}

The controlled sweeps of Sections~\ref{sec:where} through \ref{sec:how} vary one
factor at a time inside a single pipeline. This section asks what survives
outside it: across independently chosen configurations and repeated runs, across
benchmarks other than the one used throughout, and across a different model
family, adaptation regime and domain.

\subsection{Robustness across runs and configurations}
\label{sec:robustness}

Four adaptation runs were configured independently, before the controlled sweeps
existed, by different members of the project. They differ in rank, target
modules, step budget, learning rate, batch shape, weight decay and schedule
(Table~\ref{tab:configs}; full settings in Appendix~\ref{app:configs}), and each
produced its own expert selection, pruned models and evaluation under the second
pipeline of Appendix~\ref{sec:setup}.

\begin{table}[t]
\centering
\small
\caption{Four independently configured adaptation runs, second pipeline.
MMLU-Pro accuracy (\%) by experts pruned per layer, each against its own
unpruned baseline. All selections use the weight score of
Equation~\ref{eq:abs}.}
\label{tab:configs}
\begin{tabular}{lccccccc}
\toprule
& & & & \multicolumn{4}{c}{Experts pruned} \\
\cmidrule(lr){5-8}
Run & Rank & Target modules & Steps & 1 & 2 & 3 & 4 \\
\midrule
A & 16 & all linear & 1000 & 41.81 & 38.40 & 32.73 & 26.62 \\
B & 4 & router + experts & 500 & 41.03 & 36.46 & 32.35 & 26.42 \\
C & 4 & all linear & 200 & 40.53 & 36.46 & 29.87 & 24.60 \\
D & 4 & router + attn + $\mathrm{gate\_proj}$ & 100 & 39.91 & 35.68 & 29.75 & 24.39 \\
\bottomrule
\end{tabular}
\end{table}

\textbf{Configuration moves the result by about as much as rerunning does.} The
four runs span 1.90 points at one expert pruned, 2.72 at two, 2.98 at three and
2.23 at four. The largest difference we observe between repeated runs of a single
configuration is 2.24 points (below). Choosing a different rank, module set,
budget and schedule therefore displaces the outcome by roughly the same amount as
holding everything fixed and rerunning, and every one of the four curves decays
gradually, with a maximum deviation from a straight line between baseline and
endpoint of 0.81 to 2.41 points and no collapse anywhere. The shape of the result
is a property of the criterion, not of a configuration.

\textbf{The controlled trends reappear as correlations.} Ranked by final
accuracy, the four runs order exactly by adaptation steps: 1000, 500, 200 and 100
steps give 26.62\%, 26.42\%, 24.60\% and 24.39\%. The two longer runs also lose
less from their own baselines (18.7 and 18.9 points at four pruned) than the two
shorter ones (21.0 and 21.0). The best run combines the highest rank with the
longest budget, consistent with the monotone rank effect of
Appendix~\ref{sec:rank}, though since these runs vary several factors at once we
read the ordering as corroboration rather than as an independent measurement.

\begin{table}[t]
\centering
\small
\caption{Replication of the rank sweep. Three configurations were run twice
during the project under nominally identical settings; rank 8 was run once.
Cells give both runs as first\,/\,second. MMLU-Pro (\%).}
\label{tab:replication}
\begin{tabular}{lcccc}
\toprule
& \multicolumn{4}{c}{Experts pruned per layer} \\
\cmidrule(lr){2-5}
Rank & 1 & 2 & 3 & 4 \\
\midrule
2 & 39.13\,/\,38.91 & 33.09\,/\,33.17 & 25.97\,/\,27.65 & 23.46\,/\,23.42 \\
4 & 40.11\,/\,39.58 & 33.27\,/\,34.52 & 32.35\,/\,30.11 & 27.21\,/\,25.87 \\
16 & 41.12\,/\,40.64 & 36.96\,/\,36.34 & 32.66\,/\,32.21 & 28.94\,/\,28.76 \\
\bottomrule
\end{tabular}
\end{table}

\textbf{What replication preserves and what it does not.} Across the twelve
paired accuracies of Table~\ref{tab:replication}, repeated runs differ by 0.76
points on average, 0.51 at the median, and 2.24 at most; these are the figures
behind the resolution adopted in Appendix~\ref{sec:setup}. Within that noise, the
claims of Appendix~\ref{sec:rank} are exactly the ones that survive. Rank 16 is
best and rank 2 worst in both runs at every pruning level. What does not survive
is finer: in the first run, rank 4 exceeds rank 8 at three experts pruned by 0.98
points, the one inversion of adjacent ranks anywhere in either run, and its size
is characteristic of the noise. Replication thus draws the same line the
resolution analysis drew: endpoint orderings are stable, adjacent orderings are
not guaranteed, and single comparisons below two points should not be read.

\subsection{Across benchmarks}
\label{sec:benchmarks}

\begin{figure}[t]
\centering
\includegraphics[width=0.82\linewidth]{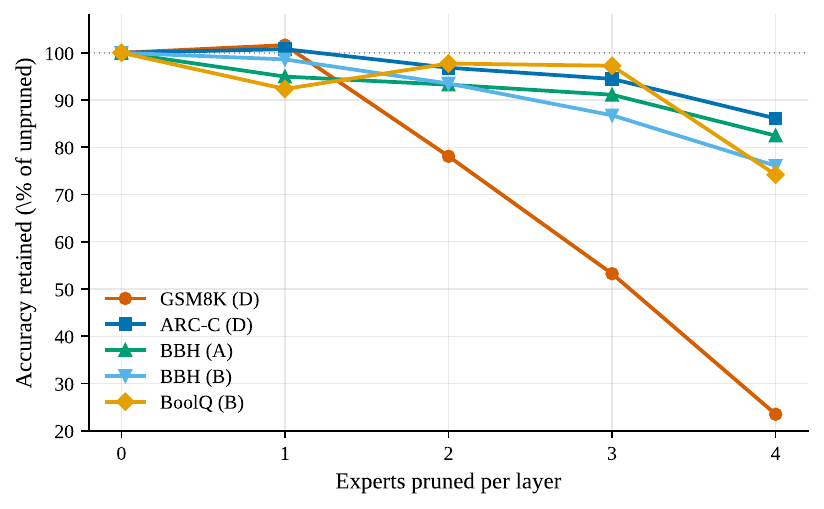}
\caption{Accuracy retained relative to each benchmark's own unpruned score, for
pruned models from the selections of Table~\ref{tab:configs}
(Table~\ref{tab:benchmarks}). Plotting retention rather than raw accuracy makes
the task-dependent \emph{shape} of the decay comparable across benchmarks with
different baselines: ARC-Challenge and BIG-Bench Hard fall gently, GSM8K is flat
then collapses, and BoolQ is non-monotonic. The dotted line marks the unpruned
level.}
\label{fig:benchmarks}
\end{figure}

MMLU-Pro has carried every result so far. Table~\ref{tab:benchmarks} evaluates
pruned models from the selections of Table~\ref{tab:configs} on four further
benchmarks; each suite is evaluated on one selection, and BIG-Bench Hard on two.
Figure~\ref{fig:benchmarks} shows the retention curves.

\begin{table}[t]
\centering
\small
\caption{Accuracy (\%) by experts pruned, on benchmarks beyond MMLU-Pro. Each
row uses pruned models from one selection of Table~\ref{tab:configs} (BBH from
two). Reported standard errors are 1.4 points for GSM8K and ARC.}
\label{tab:benchmarks}
\begin{tabular}{lcccccc}
\toprule
& & \multicolumn{5}{c}{Experts pruned per layer} \\
\cmidrule(lr){3-7}
Benchmark & Selection & 0 & 1 & 2 & 3 & 4 \\
\midrule
GSM8K & D & 51.25 & 52.08 & 40.03 & 27.29 & 12.05 \\
ARC-Challenge & D & 54.10 & 54.52 & 52.39 & 51.11 & 46.59 \\
BIG-Bench Hard & A & 43.70 & 41.50 & 40.75 & 39.82 & 36.04 \\
BIG-Bench Hard & B & 43.70 & 43.09 & 40.86 & 37.92 & 33.25 \\
BoolQ & B & 83.70 & 77.30 & 81.80 & 81.40 & 62.10 \\
\bottomrule
\end{tabular}
\end{table}

\textbf{The shape of the curve is task-dependent.} Quasi-linear decay, the
signature result on MMLU-Pro, is not universal. ARC-Challenge and BIG-Bench Hard
decline gently, losing 7.5 and 7.7 to 10.5 points across the full range. GSM8K
is flat to one expert pruned and then collapses, from 52.08\% to 12.05\%, the
steepest fall we observe anywhere on Mixtral. BoolQ is non-monotonic, dipping 6.4
points at one expert pruned, returning to within 2.3 points of baseline at two
and three, then dropping 19 points at four. For the same selection D, MMLU-Pro loses
21.0 points end to end, placing the primary benchmark mid-pack rather than at
either extreme. The selections preserve broad capability at substantial
compression while multi-step generative arithmetic degrades first and fastest,
which is consistent with such computation depending on a small set of experts
that mid-ranked removals eventually reach.

\textbf{The signal is conditioned on the adaptation task.} These selections were
produced by adapting on a generic text-generation corpus that is disjoint from
every benchmark evaluated here, so nothing in the procedure was tuned to the
tests. Read in that light, the table shows both faces of task conditioning. A
generic corpus recruits the experts that generic instruction-following uses, and
pruning by that signal preserves broad capabilities in proportion to how broadly
they are exercised; a capability the corpus exercises lightly, such as chained
arithmetic, receives no protection, and its experts sit in the middle of the
ranking where deeper pruning reaches them. The practical lever this implies is
that the adaptation corpus is a control surface: a practitioner who needs
arithmetic preserved should adapt on arithmetic. The lever is not specific to
our criterion; \citet{bai2025diep} report that switching a pruning method's
calibration data from generic text to mathematics lifts GSM8K accuracy at half
the experts by nine points or more, for their method and for an enumeration
baseline alike. Appendix~\ref{sec:qwen} evaluates exactly
that regime for router sensitivity, with the adaptation domain matched to the
evaluation domain.

\textbf{Flat is not improved.} Two curves tick upward at one expert pruned,
GSM8K by 0.83 points and ARC by 0.42. Both movements are inside the reported
standard errors of 1.4 points, so we read them as flat: removing the
lowest-sensitivity expert from every layer leaves these capabilities measurably
unchanged, which is itself the behavior the criterion promises at its first
increment.

\subsection{Across models and domains}
\label{sec:qwen}

\begin{figure}[t]
\centering
\includegraphics[width=0.8\linewidth]{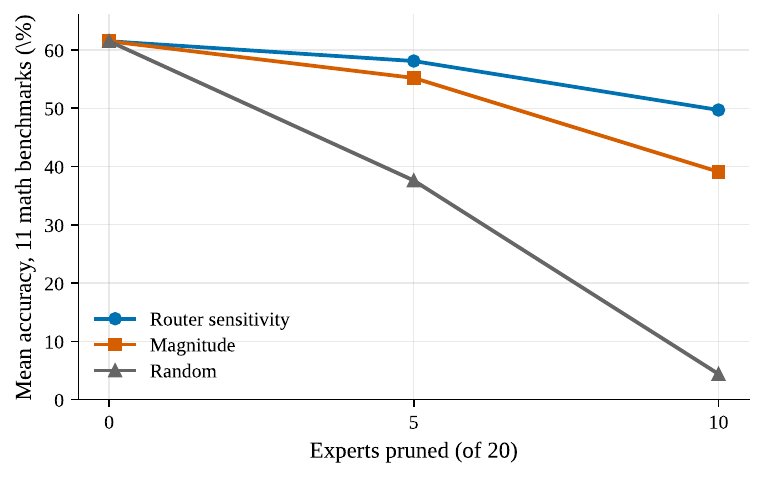}
\caption{Qwen1.5-MoE fully fine-tuned on mathematical reasoning: mean accuracy
over eleven mathematics benchmarks as experts are removed
(Table~\ref{tab:qwen}). Router sensitivity degrades gradually while random
selection collapses to single digits with half the experts removed; magnitude
selection lies between them.}
\label{fig:qwen}
\end{figure}

The remaining question is whether any of this is specific to Mixtral, to
parameter-efficient adaptation, or to general instruction data. The setting of
Appendix~\ref{sec:setup} changes all three at once: Qwen1.5-MoE with 20 experts
per layer, full fine-tuning of every parameter, and a domain-matched corpus in
place of a generic one, with adaptation on mathematical reasoning and evaluation
on eleven mathematics benchmarks, experts removed five and ten at a time, half
the experts at the deepest cut (Figure~\ref{fig:qwen}).

\begin{table}[t]
\centering
\small
\caption{Qwen1.5-MoE fully fine-tuned on mathematical reasoning: mean accuracy
(\%) over eleven mathematics benchmarks, with one representative benchmark shown.}
\label{tab:qwen}
\begin{tabular}{lccc}
\toprule
& \multicolumn{3}{c}{Experts pruned (of 20)} \\
\cmidrule(lr){2-4}
Selection criterion & 0 & 5 & 10 \\
\midrule
\multicolumn{4}{l}{\emph{Mean over eleven benchmarks}} \\
Router sensitivity & 61.5 & 58.1 & 49.7 \\
Magnitude & 61.5 & 55.2 & 39.1 \\
Random & 61.5 & 37.6 & 4.4 \\
\midrule
\multicolumn{4}{l}{\emph{MMLU College Mathematics}} \\
Router sensitivity & 62.4 & 59.0 & 50.8 \\
Magnitude & 62.4 & 56.2 & 39.8 \\
Random & 62.4 & 38.2 & 4.5 \\
\bottomrule
\end{tabular}
\end{table}

\textbf{The criterion transfers, and the alternatives do not.} With half the
experts removed, router sensitivity retains 49.7\% on average against 39.1\% for
magnitude selection, while random selection falls to 4.4\%, single-digit
accuracy on generative mathematics, a model that has effectively stopped
producing correct answers. The collapse is not an artifact of our setting:
\citet{bai2025diep} report random expert removal at the same ratio scoring
0.68\% on GSM8K. The gap to magnitude at ten experts pruned, more than
ten points on the mean and present in every one of the eleven benchmarks, is far
beyond any noise scale in this paper. The signal read from a fully fine-tuned
router on domain data behaves as it did when read from a rank-8 adapter on
general instructions.

\textbf{The location result recurs.} In preliminary Qwen adaptations,
router-only LoRA reached 29.31\%
MMLU-Pro against 28.13\% for LoRA on all modules. The gap of 1.18 points sits within the
measurement resolution; its direction repeats, on a second model family, the
central finding of Appendix~\ref{sec:where}.

\section{Positioning against one-shot pruning methods}
\label{sec:compare}

\textbf{This work evaluates what the literature does not.} To our knowledge, no
published study reports MMLU-Pro for expert-pruned Mixtral. REAP evaluates its
Mixtral compressions on ARC, BoolQ, HellaSwag, MMLU, OpenBookQA, RTE and
WinoGrande \citep{lasby2025reap}; DiEP on MMLU, OpenBookQA, BoolQ and RTE
\citep{bai2025diep}; trajectory-based selection on MMLU, HellaSwag, WinoGrande,
ARC, GSM8K and MedQA \citep{yang2025pathfinder}; the task-agnostic line on MMLU,
BoolQ, OpenBookQA and RTE \citep{jaiswal2025finding}. Standard MMLU is the
common denominator of that literature, and its four-option format leaves a
degraded model substantial guessing headroom. None of the five suites includes BIG-Bench Hard
either. Every MMLU-Pro and BBH number in Sections~\ref{sec:where} through
\ref{sec:generalize} is therefore the first characterization of expert-pruned
Mixtral on these benchmarks, and the quasi-linear decay established on MMLU-Pro
across five compression points is a property no previous study, reporting two
points on four-option MMLU, was positioned to observe.

\textbf{On the benchmarks the literature does report, the signal is
competitive.} Table~\ref{tab:sota} compares retention, each pruned score
normalized by its own study's unpruned baseline, on the three benchmarks where
our evaluations and published ones overlap. At 25\% compression the signal sits
inside the published band everywhere: BoolQ retention of 97.7\% against 98.3\%
for REAP and 99.1\% for DiEP, ARC retention of 96.8\% above both reported
values, and GSM8K retention of 78.1\%, above AIMER's 73.4\% on the same Mixtral
checkpoint under a different harness \citep{liu2026aimer}. These are
methods built on full activation statistics, differentiable search over expert
subsets, and global trajectory planning; the row they bracket comes from rank-4
LoRA selections of Table~\ref{tab:configs}, and in its leanest router-only form
the signal trains 0.002\% of parameters.

\begin{table}[t]
\centering
\small
\caption{Retention (\%) of each study's own unpruned baseline after expert
pruning on Mixtral-8$\times$7B-Instruct, at 25\% and 50\% expert sparsity, on
the benchmarks where evaluations overlap. Published figures are computed within
their own harness and settings; AIMER's and REAP's ARC values use normalized accuracy, ours
ordinary accuracy; DiEP and trajectory selection allocate experts nonuniformly
across layers, whereas we remove uniformly; the trajectory GSM8K result is
calibrated on task data, whereas our selections are conditioned on a generic
corpus. Dashes mark benchmarks a study does not report.}
\label{tab:sota}
\begin{tabular}{lcccccc}
\toprule
& \multicolumn{2}{c}{BoolQ} & \multicolumn{2}{c}{ARC-Challenge} & \multicolumn{2}{c}{GSM8K} \\
\cmidrule(lr){2-3} \cmidrule(lr){4-5} \cmidrule(lr){6-7}
Selection criterion & 25\% & 50\% & 25\% & 50\% & 25\% & 50\% \\
\midrule
Router sensitivity (this work) & 97.7 & 74.2 & 96.8 & 86.1 & 78.1 & 23.5 \\
AIMER v3 \citep{liu2026aimer} & 98.8 & 97.7 & 92.9 & 87.4 & 73.4 & 31.6 \\
REAP v3 \citep{lasby2025reap} & 98.3 & 94.8 & 92.3 & 84.0 & -- & -- \\
DiEP \citep{bai2025diep} & 99.1 & 96.7 & -- & -- & -- & -- \\
MoE Pathfinder \citep{yang2025pathfinder} & -- & -- & 94.4 & 85.0 & 83.9 & 46.3 \\
\bottomrule
\end{tabular}
\end{table}

\textbf{At 50\% the ordering is task-dependent and tracks conditioning.} ARC
retention of 86.1\% exceeds REAP's 84.0\% and matches trajectory selection's
85.0\%. BoolQ and GSM8K fall below the published values. The GSM8K column
is the conditioning effect of Appendix~\ref{sec:benchmarks} measured across
papers: the published figure is task-calibrated, ours is generically
conditioned, and DiEP's own ablation prices that difference at nine points or
more of GSM8K accuracy at this compression \citep{bai2025diep}.

\textbf{Competitive at moderate compression, and the economics run in one
direction.} The value of the criterion is a combination that the table cannot show
in one column. On overlapping benchmarks at moderate compression it is
competitive with methods that consume full activation statistics or learned
allocations, and the economics run in one direction twice over. Against the
full fine-tuning the criterion originally required, the router-only signal
trains 0.002\% of the parameters. Against calibration statistics, which are
always a dedicated expense, the common deployment path already includes
parameter-efficient adaptation, so the ranking arrives as a byproduct of work
performed anyway, at zero marginal cost. It
uniquely covers the benchmark class the literature avoids, where it establishes
predictable quasi-linear decay. And it is the one criterion whose knobs this
paper has mapped: Sections~\ref{sec:where} through \ref{sec:how} identify
which adaptation choices sharpen the ranking, and
Appendix~\ref{sec:benchmarks} shows the adaptation corpus steers which
capabilities pruning protects, a steering that external ablations confirm for
calibration data as well \citep{bai2025diep} but that no other criterion
arrives with characterized.

\section{Deployment efficiency}
\label{sec:deploy}

Accuracy retention is only half of the pruning case; the other half is what
removal buys at inference time. We measure GPU memory and per-token generation
latency with Optimum-Benchmark \citep{moutawwakil2023optimum} on a single A100,
with base weights quantized to 4 bits as in adaptation, for the unpruned model
and the four router-sensitivity pruned variants of Appendix~\ref{sec:where}.

\begin{table}[ht]
\centering
\small
\caption{Inference cost by experts retained, measured with Optimum-Benchmark on
one A100 under 4-bit weights. Memory is in GB and is a 4-bit measurement, so the
unpruned figure of 24.2\,GB is unrelated to the 24.2B \emph{parameter} count of
the four-experts-pruned model elsewhere in the paper. Relative throughput is the
reciprocal of per-token latency and therefore describes single-stream decoding.}
\label{tab:deploy}
\begin{tabular}{lcccc}
\toprule
Model & Experts & Memory (GB) & Latency/token (ms) & Rel.\ throughput \\
\midrule
Unpruned & 8 & 24.2 & 40.3 & 100\% \\
1 pruned & 7 & 21.3 & 35.8 & 112.6\% \\
2 pruned & 6 & 18.1 & 32.4 & 124.4\% \\
3 pruned & 5 & 15.4 & 28.7 & 140.4\% \\
4 pruned & 4 & 12.3 & 25.2 & 159.9\% \\
\bottomrule
\end{tabular}
\end{table}

\textbf{Costs fall near-linearly in experts removed.} Memory decreases by 2.7 to
3.2\,GB per expert removed, 2.97 on average, against the 2.8\,GB that one
expert's 5.6B parameters occupy at 4 bits; the small surplus is the removed rows'
share of activation and dequantization overhead. Latency decreases by 3.4 to
4.5\,ms per expert. Neither series deviates from a straight line between its
endpoints by more than 0.15\,GB or 0.73\,ms. At the deepest cut the model uses
49\% less memory and generates each token 37\% faster, and single-stream
throughput rises 60\%.

\textbf{Under this implementation, we attribute the latency gain to memory traffic rather than FLOPs.} Top-2
routing is unchanged by pruning: every token activates two experts before and
after, so the arithmetic per token is constant across the rows of
Table~\ref{tab:deploy}. What shrinks in the measured implementation is the set of
distinct expert weights touched per forward pass. Across the tokens of a
sequence the router reaches most resident experts in every layer, so the weights
streamed from memory scale with the number retained. The observed 37\% reduction
is real for this stack, but its magnitude may depend on expert dispatch,
quantization, and kernel implementation.

\textbf{Both sides of the trade are predictable.} The cost curves here are
near-linear in experts removed, and Appendix~\ref{sec:where} showed the MMLU-Pro
accuracy curve under router-sensitivity selection to be quasi-linear over the
same range. A practitioner can therefore price a compression target in advance
from the endpoints alone: on this hardware, each expert removed buys roughly
3\,GB of memory and 3.5 to 4.5\,ms per token, at a quasi-linear accuracy cost
whose slope Sections~\ref{sec:where} through \ref{sec:how} showed how to
minimize. Neither the naive selection rules, which collapse abruptly past two
experts, nor benchmarks with cliff behavior such as GSM8K offer that
predictability, which is why we regard the shape of the decay, as much as its
level, as the practically valuable property of the criterion.

\section{Discussion}
\label{sec:discussion}

\textbf{Why the router concentrates the signal.} The results of
Sections~\ref{sec:where} through \ref{sec:how} admit a single mechanism. The
gate matrix is the one place in the network where the decision of which experts
serve a token is expressed as parameters, so it is the one place where a task's
revision of expert recruitment must appear if it appears at all. Whether it
appears cleanly depends on the alternatives the optimizer has. Give the adapter
access to attention and expert weights and loss can fall by changing what
experts compute rather than which are consulted, and the gate moves less
distinctively (27.54\% against 24.42\% at four experts pruned). Give the router
adapter too little internal width and parameter budget, and it registers a
weaker task-conditioned revision, degrading the ranking monotonically as size
shrinks. Give
the adaptation freedom to inject arbitrary directions into the representation
and the gate moves for reasons unrelated to expert utility, which is where
Houlsby adapters lose to multiplicative scaling. The signal is faithful when the perturbation reaching the router is both
forced, because the task cannot be absorbed away from routing, and disciplined,
because the perturbation cannot reorganize routing for reasons unrelated to
expert utility. Every degradation we measure violates one of the two, and
Appendix~\ref{sec:how} shows the perturbation need not write to $W_g$ at all to
satisfy both.

\textbf{An instrument objective inverts training guidance.} Practical guidance
for LoRA as a training method recommends attaching it broadly, with the
feed-forward and expert layers mattering most, because that configuration best
matches full fine-tuning on task loss \citep{dettmers2023qlora,
thinkingmachines2025lora}. Our measurements agree with that guidance on its own
terms and invert it on ours: the all-modules configuration is the better way to
produce a capable model and the worse way to decide what to delete. The
divergence is principled rather than paradoxical. Training optimizes the
model's outputs; an instrument optimizes the legibility of a specific internal
change. Sections~\ref{sec:where} through \ref{sec:how} can be read as the
beginning of a configuration guide for the second objective, for which the
PEFT literature's first-objective guidance does not transfer.

\textbf{What the empirics say back to the theory.} The guarantee of
\citet{chowdhury2024provably} is stated for full fine-tuning and for the signed
norm change. Our results extend its empirical reach and, in three places, ask
questions the theory does not yet answer. First, the monotone rank effect:
scaling the adapter's internal width changes the ranking gracefully and
monotonically rather than abruptly, which invites an analysis of the criterion
under parameter-efficient adaptation. Second, perturbation geometry: a
multiplicative rescaling that never writes to $W_g$ ranks experts as well as a
direct low-rank update while the additive alternative ranks them worse under
the tested budgets, a distinction absent from the current analysis. Third, the sign: our
score is the magnitude of the norm change, the guarantee concerns its signed
value, and our sensitivity maps show gate norms that contract as well as grow
under adaptation. Whether contraction marks an expert as recruited or as
released is a well-posed question settled by re-ranking gate norms from the
adapted checkpoints at no training cost, and the answer would sharpen both the
theory and the practice.

\textbf{The module ladder has two unscored rungs.} Attention-only and
expert-only adaptation were excluded from the module sweep because they leave
$W_g$ untouched and Equation~\ref{eq:abs} identically zero. Equation~\ref{eq:logit}
scores them anyway, by reading the logits the frozen router produces over
shifted representations. Running the module ladder under the logit score would
separate two claims this paper currently cannot distinguish: that the router
must be adapted for the signal to exist, and that it must merely be observed
while the model is under task pressure. Appendix~\ref{sec:how} already shows the
second suffices for IA3; completing the ladder is the cleanest single
experiment this work leaves open.

\textbf{Calibration, and where the one-shot numbers sit.} Every result in this
paper is one-shot and uncalibrated by design, isolating the selection criterion
from any repair applied after it. \citet{hyeon2026retrainingfree} show that
distilling into the router after compression recovers substantial accuracy on
fine-grained architectures while yielding only marginal recovery on Mixtral,
whose eight experts per layer admit few routing configurations. Read together,
the two results place our Mixtral numbers near the ceiling of what router-side
repair would add, and mark the fine-grained regime, including the Qwen setting
of Appendix~\ref{sec:qwen}, as where a calibration stage is the natural
extension; recovery through retraining the surviving experts is a separate and
larger budget that our one-shot scope deliberately excludes.
Quasi-linear uncalibrated decay is the floor that stage would build on.

\textbf{Scope.} The controlled sweeps cover one model family, and a controlled head-to-head evaluation remains future work. Sweep results are single runs read
against a 2.2-point replication
resolution, with the twice-run configurations reported as ranges in
Appendix~\ref{sec:robustness} and Appendix~\ref{app:replication}. The IA3 and
Houlsby comparisons use their own training budget rather than one matched to
the LoRA sweeps. Configuration selection throughout was keyed to MMLU-Pro, and
Appendix~\ref{sec:benchmarks} quantifies how far that choice generalizes. The
signed-score question above is open, and costs only a re-ranking from the
adapted checkpoints to close.

\section{Conclusion and future work}
\label{sec:conclusion}

A provably motivated criterion for expert pruning, previously defined against
full fine-tuning, is practical at the scale of current MoE language models: a
brief parameter-efficient adaptation moves the router enough to rank experts,
the magnitude of each gate vector's norm change (or, when the router is frozen,
the shift in its logits) supplies the ranking, and one-shot removal of the
lowest-ranked experts yields models whose MMLU-Pro accuracy decays
quasi-linearly with compression while memory and latency fall near-linearly.
The signal concentrates where the decision it measures is expressed. Confining
adaptation to the router sharpens it, capacity spent there improves it
monotonically, and the tested constrained perturbation preserves it better than
the additive alternative under their shared budget. The ranking survives changes of configuration,
benchmark, model family, adaptation regime and domain, and on the benchmarks it
shares with recent one-shot pruning methods it is competitive at moderate
compression from a fraction of their signal cost.

For a practitioner the recipe is short. Attach LoRA to the router alone at
moderate rank, adapt briefly on a corpus that exercises the capabilities the
pruned model must keep, rank by norm change, and remove experts uniformly per
layer to the target budget, stopping earlier for capabilities that cliff. The
corpus is a control surface and the rank a fidelity knob, both cost a
negligible fraction of the model they compress, and where deployment already
includes parameter-efficient adaptation the ranking arrives as a byproduct, at
no additional training cost.

Several directions follow directly. Owing to computational constraints, the
controlled sweeps concentrate on a single coarse-grained family; validating the
configuration findings across MoE families, from fine-grained architectures
with many small experts and shared-expert designs to larger Mixtral variants,
is the most direct extension, and the Qwen results of Appendix~\ref{sec:qwen} mark the path. Running a controlled evaluation of every criterion on the same checkpoint, calibration corpus, and sample budget would extend the published-method comparison in Appendix~\ref{sec:compare}. Such an evaluation should remove the same number of experts per layer, apply no post-pruning repair, and use identical LightEval prompts and configurations across MMLU-Pro, GSM8K, ARC-Challenge, BIG-Bench Hard, and BoolQ. Completing the module ladder under the logit score separates adaptation of the router from observation of it, and re-ranking stored gate norms answers whether the sign of the change carries information. Beyond the one-shot scope, iterative pruning
with interleaved healing and a router-calibration stage are the natural recovery mechanisms, with the most to gain in the fine-grained regime. On the theoretical side, the monotone rank effect and the role of perturbation geometry are regularities the current guarantee does not predict, and
extending the analysis to constrained adaptation would close the loop this paper opens. Each direction sharpens the same instrument this work set out to
build.

\clearpage
\section{Artifacts and full hyperparameter configurations}
\label{app:configs}

Code for adaptation, sensitivity scoring, pruning and evaluation is available
through the GitHub artifact:
\begin{flushleft}
\url{https://github.com/ianKa1/MoE_pruning}
\end{flushleft}
The training data are available through the training-dataset
repository:
\begin{flushleft}
\url{https://huggingface.co/datasets/Na0s/sft-ready-Text-Generation-Augmented-Data}
\end{flushleft}
Model checkpoints and artifacts can also be found here:
\begin{flushleft}
\small
\url{https://huggingface.co/Na0s/Mixtral-8x7B-Instruct-v0.1-LoRA-on-Gates}\\
\url{https://huggingface.co/Na0s/Mixtral-8x7B-Instruct-v0.1-exhaustive-LoRA}\\
\url{https://huggingface.co/Na0s/Mixtral-8x7B-v0.1-instruct-pruned-random-1-experts}\\
\url{https://huggingface.co/Na0s/Mixtral-8x7B-v0.1-instruct-pruned-random-2-experts}\\
\url{https://huggingface.co/Na0s/Mixtral-8x7B-v0.1-instruct-pruned-random-3-experts}\\
\url{https://huggingface.co/Na0s/Mixtral-8x7B-v0.1-instruct-pruned-random-4-experts}\\
\url{https://huggingface.co/Na0s/Mixtral-8x7B-v0.1-instruct-l2-norm-post-Gates-LoRA-SFT-pruned-1-expert}\\
\url{https://huggingface.co/Na0s/Mixtral-8x7B-v0.1-instruct-l2-norm-post-Gates-LoRA-SFT-pruned-2-experts}\\
\url{https://huggingface.co/Na0s/Mixtral-8x7B-v0.1-instruct-l2-norm-post-Gates-LoRA-SFT-pruned-3-experts}\\
\url{https://huggingface.co/Na0s/Mixtral-8x7B-v0.1-instruct-l2-norm-post-Gates-LoRA-SFT-pruned-4-experts}\\
\url{https://huggingface.co/Na0s/Mixtral-8x7B-Instruct-v0.1-exhaustive-LoRA-SFT-pruned-1-expert}\\
\url{https://huggingface.co/Na0s/Mixtral-8x7B-Instruct-v0.1-exhaustive-LoRA-SFT-pruned-2-experts}\\
\url{https://huggingface.co/Na0s/Mixtral-8x7B-Instruct-v0.1-exhaustive-LoRA-SFT-pruned-3-experts}\\
\url{https://huggingface.co/Na0s/Mixtral-8x7B-Instruct-v0.1-exhaustive-LoRA-SFT-pruned-4-experts}
\end{flushleft}

Table~\ref{tab:appconfigs} lists the settings governing each adaptation run
reported in the paper; the multiplicative and additive-adapter budgets and the
two full fine-tuning references are given in Appendix~\ref{sec:setup}. All runs
use the 8-bit AdamW optimizer.

\begin{table}[t]
\centering
\small
\caption{Adaptation configurations. Canonical denotes the reference run of the
controlled sweeps; A through D are the independently configured runs of
Appendix~\ref{sec:robustness}. Batch is per-device batch size $\times$ gradient
accumulation.}
\label{tab:appconfigs}
\begin{tabular}{lccccc}
\toprule
& Canonical & A & B & C & D \\
\midrule
Rank $r$ / $\alpha$ & 8\,/\,8 & 16\,/\,16 & 4\,/\,4 & 4\,/\,4 & 4\,/\,4 \\
Target modules & router & all linear & router + experts & all linear & router + attn + $\mathrm{gate\_proj}$ \\
Steps & 500 & 1000 & 500 & 200 & 100 \\
Learning rate & $2\times10^{-4}$ & $2\times10^{-4}$ & $1\times10^{-5}$ & $2\times10^{-4}$ & $1.5\times10^{-4}$ \\
Batch & $1\times20$ & $1\times16$ & $1\times16$ & $8\times2$ & $1\times4$ \\
Warmup & 5 steps & 5 steps & ratio 0.05 & 5 steps & ratio 0.03 \\
Weight decay & 0.01 & 0.01 & 0.03 & 0.01 & 0.04 \\
Schedule & cosine & cosine & cosine & cosine & cosine, restarts \\
Max grad norm & -- & 0.3 & 0.3 & 0.3 & 0.3 \\
LoRA dropout & 0.1 & 0.1 & 0.1 & 0.1 & 0.1 \\
\bottomrule
\end{tabular}
\end{table}

\section{Replication tables}
\label{app:replication}

Table~\ref{tab:replication} in the main text reports the three rank
configurations run twice under nominally identical settings; their twelve
paired accuracies differ by 0.76 points on average, 0.51 at the median and 2.24
at most, which fixes the measurement resolution of Appendix~\ref{sec:setup}. Two
module-sweep configurations also exist in both project phases under differing
names, and Table~\ref{tab:appmodrepl} reports them under the mapping that
identifies the phase-one names Router+Attention and Router+Experts with the
phase-two target sets gate$+q,k,v,o$ and gate$+$MLP. Their eight paired
accuracies differ by 0.43 points on average and 0.97 at most, tightening rather
than loosening the resolution; the main text conservatively counts only the
three rank configurations, whose identity across phases is unambiguous.

\begin{table}[t]
\centering
\small
\caption{Module-sweep configurations run in both project phases, under the
name mapping stated in the text. Cells give both runs as
first\,/\,second. MMLU-Pro (\%).}
\label{tab:appmodrepl}
\begin{tabular}{lcccc}
\toprule
& \multicolumn{4}{c}{Experts pruned per layer} \\
\cmidrule(lr){2-5}
Target modules & 1 & 2 & 3 & 4 \\
\midrule
gate + $q,k,v,o$ & 40.13\,/\,39.94 & 35.68\,/\,34.98 & 31.11\,/\,30.40 & 26.90\,/\,25.93 \\
gate + MLP & 39.96\,/\,39.82 & 35.03\,/\,34.85 & 30.33\,/\,29.97 & 25.19\,/\,24.97 \\
\bottomrule
\end{tabular}
\end{table}

\section{Router sensitivity across configurations}
\label{app:heatmaps}

Figure~\ref{fig:heatmap} shows the sensitivity map for the all-module rank-16
run; the maps from the other independently configured runs of
Appendix~\ref{sec:robustness} share its qualitative structure, differing mainly
in overall scale with the adaptation budget. In every case sensitivity is far
from uniform, most cells sitting near their layer median while a few
expert-layer pairs are an order of magnitude larger, and in no case is the mass
concentrated in a single expert column or a single band of layers, which is what
motivates per-layer ranking rather than a global threshold
(Appendix~\ref{sec:prune}). The signed variant of these maps shows gate norms
that contract as well as grow under adaptation, the observation behind the sign
question of Appendix~\ref{sec:discussion}.

\section{Sensitivity scoring code}
\label{app:code}

The two scores of Appendix~\ref{sec:scores}, written over a generic accessor
because the router attribute path differs by model family
(\texttt{block\_sparse\_moe.gate} in HuggingFace Mixtral, \texttt{mlp.gate} in
Qwen-MoE).

{\footnotesize
\begin{verbatim}
def router(layer):
    # Mixtral: layer.block_sparse_moe.gate ; Qwen-MoE: layer.mlp.gate
    return layer.block_sparse_moe.gate.weight      # [num_experts, hidden]

def weight_scores(base, adapted):
    # s_e = | ||w_e^(T)|| - ||w_e^(0)|| |  per layer   (main-text Eq. 1)
    scores = []
    for lb, la in zip(base.model.layers, adapted.model.layers):
        wb = router(lb)
        wa = router(la)
        n0 = torch.linalg.norm(wb, ord=2, dim=1)
        nT = torch.linalg.norm(wa, ord=2, dim=1)
        scores.append((nT - n0).abs())
    return scores          # rank ascending, prune the smallest per layer
\end{verbatim}
}

{\footnotesize
\begin{verbatim}
def logit_scores(base, adapted, batch):
    # s_e = ||dL[:, e]||_2 / sqrt(N)   over held-out tokens   (main-text Eq. 2)
    # Shape: [layers, N, num_experts]
    L0 = collect_router_logits(base, batch)
    LT = collect_router_logits(adapted, batch)
    d = LT - L0
    return d.norm(dim=1) / math.sqrt(d.shape[1])  # [layers, num_experts]
\end{verbatim}
}
